\documentclass[10pt,twocolumn,letterpaper]{article}

\usepackage{cvpr}
\usepackage{times}
\usepackage{epsfig}
\usepackage{graphicx}
\usepackage{amsmath}
\usepackage{amssymb}

\usepackage[breaklinks=true,bookmarks=false]{hyperref}

\cvprfinalcopy 

\begin{document}

\title{\LaTeX\ Detection of Premature Ventricular Contractions Using Densely Connected  Deep Convolutional Neural Network with Spatial Pyramid Pooling Layer}

\author{Jianning Li\\
{\tt\small }
}

\maketitle

\begin{abstract}
Premature ventricular contraction(PVC) is a type of premature ectopic beat originating from the ventricles. Automatic method for accurate and robust detection of PVC  is highly clinically desired.Currently, most of these methods are developed and tested using the same database divided into training and testing set and their generalization performance across databases has not been fully validated. In this paper, a  method based on densely connected convolutional neural network and spatial pyramid pooling is proposed for PVC detection which can take arbitrarily-sized  QRS complexes as input both in training and testing. With a much more straightforward architecture,the proposed network achieves comparable results to current state-of-the-art deep learning based method with regard to accuracy,sensitivity and specificity by training and testing using the MIT-BIH arrhythmia database as benchmark.Besides the benchmark database,QRS complexes are extracted from four more open databases namely the St-Petersburg Institute of Cardiological Technics 12-lead Arrhythmia Database,The MIT-BIH Normal Sinus Rhythm Database,The MIT-BIH Long Term Database and European ST-T Database. The extracted QRS complexes are different in length and sampling rate among the five databases.Cross-database training and testing is also experimented achieving a 0.9943 overall accuracy with  0.9819  sensitivity and 0.9952 specificity  demonstrating the advantage of  using multiple databases for training over  using only a single database.The network also achieves satisfactory scores on the other four databases showing good generalization capability. 
\textbf{Keywords}: Premeture ventricular contraction,detection, denseNet,spatial pyramid pooling,deep learning
\end{abstract}

\section{Introduction}

Frequent premature ventricular contraction(PVC) is often associated with organic heart diseases and should be treated medically or surgically if it bothers life quality\cite{Frequent} while sporadic PVC can happen to most healthy population and is usually considered benign if it does not trigger severe ventricular arrhythmia such as supraventricular tachycardia and ventricular fibrillation.The two types of PVC are usually treated differently and the classification is based on how many times it occurs in a long-term ECG recording. Therefore,it’s clinically significant to develop accurate and robust methods to detect PVC automatically.
Up until now, various methods for PVC detection have been developed with the aim of  reducing doctors’ workload and these  methods can be categorized into two classes.First,hand-craft features combining a classifier.These kind of methods are often seen in earlier published works that utilize hand-craft ECG features and a classifier to distinguish between PVC  and non-PVC.The hand-craft features include morphology\cite{Hadia2017Morphology}, wavelet\cite{Jung2017Detection,Chang,Sayadi2010Robust,Wisana2012Identification,Kumar} and temporal domain\cite{Manikandan2015Robust} etc. Some works also studied methods for the selection of hand-craft features to improve  detection performance \cite{Jekova2005Pattern,Feature,Nuryani2014Premature}.Various classifiers are used in the studies of PVC detection including  artificial neural network\cite{Lim2009Finding,Nugroho2016Premature,Zhou2003Automatic},support vector machine\cite{Alajlan} as well as clustering\cite{V2008Comparison}.Second, deep convolutional neural network (CNN)is showing advantages over traditional methods by providing a way of  learning highly discriminative features automatically. Many works have studied its application in detecting abnormal heart beats including atrial fibrillation\cite{Bollepalli2017Atrial,Smolen2017Atrial,Xia2017Atrial,Atrial,Rubin2017Densely} and PVC\cite{Zhou2017Premature,Yang2015A,Zarei} or other arrhythmia\cite{Rajpurkar2017Cardiologist}. Currently,\cite{Zhou2017Premature} achieves state-of-the-art results for PVC detection using a architecture combining multiple one-dimensional CNN and LSTM.However,all of the methods mentioned above are developed and tested using  the same database and  their cross-database generalization capability  has not been fully validated.In clinic,however,the sampling rate of ECG data can be different from different devices.In such cases,it’s natural to think of training several networks for each specific ECG data which is time-consuming and sometimes impossible when the ECG data are limited.Therefore,it’s necessary to develop a generalized method  that can maintain good performance across ECG data of varied sampling rates.In our study ,we propose a method based on densely connected  convolutional neural network \cite{Huang2016Densely}and spatial pyramid pooling \cite{Kaiming2014Spatial} for automatic PVC detection.The proposed network can take as input QRS complexes of arbitrary length and can be trained using multiple ECG databases with different sampling rates.Its cross-database generalization capability is verified on five open databases  namely the MIT-BIH arrhythmia database,St-Petersburg Institute of Cardiological Technics 12-lead Arrhythmia Database,The MIT-BIH Normal Sinus Rhythm Database,The MIT-BIH Long Term Database and European ST-T Database.The performance on the MIT-BIH arrhythmia database which is a commonly used benchmark for arrhythmia detection is comparable to current state-of-the-art deep learning based  method and our proposed network is much less complicated and easier to implement.

\section{Databases}
Five open ECG databases from PhysioNet\cite{Goldberger2000PhysioBank} are involved in our study namely the MIT-BIH arrhythmia database(DS1)\cite{Moody2002The},St-Petersburg Institute of Cardiological Technics 12-lead Arrhythmia Database(DS2)\cite{Goldberger2000PhysioBank},MIT-BIH Normal Sinus Rhythm Database(DS3),MIT-BIH Long Term Database(DS4)and European ST-T Database(DS5)\cite{Taddei1992The}.Half of the recordings in each database are used for training and the other half for testing. QRS complexes are selected  from these recordings and  categorized to PVC and non-PVC based on the location of R-peaks and beat annotations provided along with the databases.For DS1,the selection of  training set and testing set is the same as that of \cite{Zhou2017Premature,Zarei} for comparison purpose.For other databases,equal number of  non-PVC beats are selected from each long-term recording and all PVC  beats are considered.The length of  the extracted QRS complexes is different across databases due to different sampling rates.According to published literature\cite{Hadia2017Morphology,Zarei},about 150 sampling points are selected around  R-peak to represent the QRS complex(50 and 99 sampling points before and after R-peak respectively ) for DS1 in our study.For DS2-DS5, the number of sampling points l$_i $ for the QRS complex  is  decided by the following equation:

\begin{equation}
{\l_i=l_0\frac{f_i}{f_0}} 
\end{equation}
where l$_0$ =150 is the the length of QRS complex from DS1 and f$_0 $ is the sampling rate of DS1.$f_i(i=2,3,4,5)$is the sampling rate for DS2-DS5.

Table 1 shows the details of the extracted QRS complex from DS1-DS5.Altogether,the training set of the  mixed database contains  32569 PVC beats and 139015 non-PVC beats.The testing set contains 62200 and 122771 beats for PVC and non-PVC respectively.Figure 1 illustrates the QRS wave morphology of PVC and non-PVC.

\begin{table*}[h]
 \caption{Mixed Database of QRS Complexes(SR=Sampling Rate)}
 \label{sample-table}
 \centering
 \begin{tabular}{lllcccc}
   \hline
 Database  &SR   &Length  &PVC(train)        &non-PVC (train)         &PVC (test)       &non-PVC(test) \\
   \hline
DS1      &360    &150           & 3788          &47213               &3220         &46478   \\
DS2      &257   & 108          &  8467          &41264               &10711       &32838  \\
DS3     & 128    & 52            &  21             &5000                &    5          &  1000  \\
DS4      &128  &  52             &19867         &20000               &44224       &20000 \\
DS5      &250   & 105           & 426          &  25538               & 4040       & 22455  \\  
Total     & /      &  /              &32569      &  139015            & 62200       & 22771   \\
   \hline
 \end{tabular}
\end{table*}

Among the five databases described above,the MIT-BIH arrhythmia database is a commonly used benchmark for developing and evaluating arrhythmia detection algorithms while the other four databases are built for various purposes. For example,the European ST-T database is intended to be used for evaluating algorithms related to S-T segment and T wave.For all databases, the location of R peaks and beat-by-beat annotations are provided.The diversity of ECG data is greatly improved compared to using only a single database.On the one hand,the ECG data in these  databases are collected from more unique individuals using different devices with varied sampling rates and analog-to-digital resolution.On the other hand,more arrhythmia types can be included to the  non-PVC category using multiple databases.Note that not all  beats of these long-term  ECG recordings are considered.
Training and testing on databases of greater diversity  helps the network generalize better and the testing results can be more reliable.

\begin{figure*}[t]
\centering
\includegraphics[width=1\linewidth]{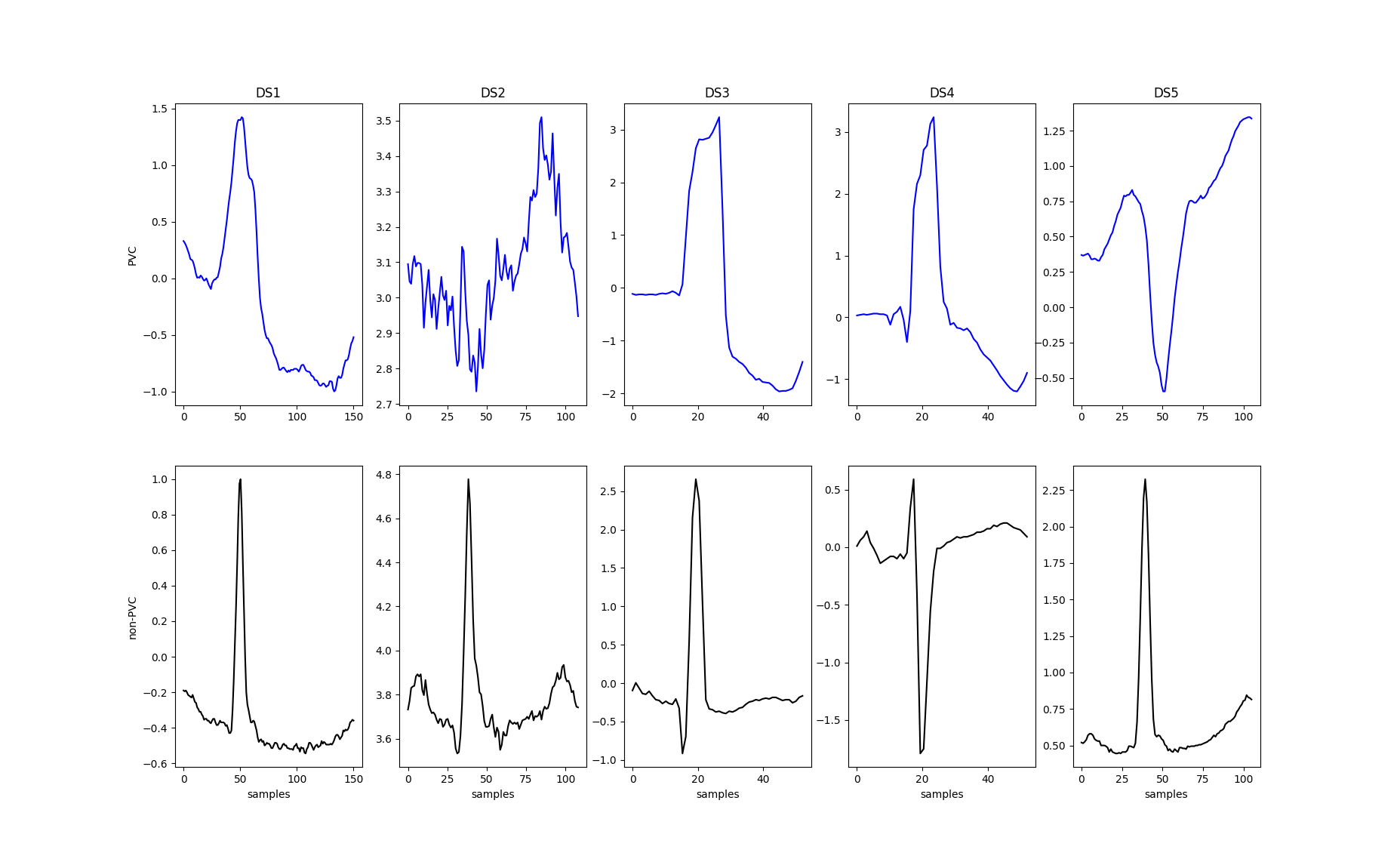}\vspace{1mm}
\caption{QRS complex extracted from long-term  ECG  in DS1-DS5, PVC(top row) and non-PVC(bottom row).}
\end{figure*}

\section{Methodology}
The proposed diagram for PVC detection is shown in Figure.2. First,the ECG signals are bandpass-filtered from 0.4Hz to 50Hz using a 4-level Butterworth filter  before the QRS complexes are extracted.Second,our network takes as input the QRS complex of arbitrary length for processing.Third, a nonlinear transformation is applied to the input to compute highly discriminative features.The nonlinear transformation is done through  three dense blocks connected by convolutional and max pooling operations.Fourth,the extracted features are vectorized  by a spatial pyramid pooling layer.Different from traditional pooling operations that produce variable sized outputs if the input size is not fixed, the spatial pyramid pooling layer accepts arbitrarily-sized inputs but outputs a fixed-length vector.Finally,the fully connected layer with sigmoid activation function produces the probability of an input belonging to PVC.Both dense blocks and spatial pyramid pooling helps reduce over-fitting and improve generalization capability which is crucial for PVC detection algorithms when it comes to real clinical application.
Details of each procedure will be specified in the following sections.

\subsection{Densely connected  block}
Proposed in \cite{Huang2016Densely},dense connections between layers alleviate the problem of vanishing-gradient as the network goes deeper.With smaller number of filters for each convolutional layer, the total number of trainable parameters in the network can maintain small even if it’s deep which helps reduce over-fitting and improve generalization capability.  
Figure.3(a) shows a dense block with 3 1-dimensional convolutional layers where all the layers are connected to each other so that  the input of each layer is the output feature map of all the preceding layers.By using dense connections among layers,the features can be made best use of.
In our study,we use three dense blocks ,each having 3,6, and 9  1-dimensional convolutional layers respectively with no bottleneck layer\footnote{A bottleneck in a dense block is a convolutional operation with kernel width set to 1 \cite{Huang2016Densely}}. The number of filters for all convolutional layers in all dense blocks is 32 and  the kernel width is set to 3. Between dense blocks are transition layers which conduct  convolutional,batch normalization and average pooling operations followed by a relu activation function . 
The size of feature maps produced within a dense block are the same because there’s no pooling operations in a dense block.The transition layer is responsible for change feature-map sizes through convolutional and pooling operations.Note that the convolutional operations in a dense block does not change the size of feature maps for layer concatenation purpose.
\subsection{Spatial pyramid pooling layer}

To address the issue of varying sizes of the extracted QRS complex due to varied sampling rate,1-dimensional spatial pyramid pooling\cite{Kaiming2014Spatial}is adopted in our network to replace the traditional max pooling operations in the last layer.Figure 3(b) shows the 1-dimensional spatial pyramid pooling layer we use for vectorizing feature maps produced by preceding convolutional layer.Assuming that the length of current input QRS is 152 and the densely connected component extracts 272 feature maps of length 19 from the input.For each feature map, max-pooling  is done on two levels of  feature pyramid.First, pooling on the entire feature map which produces one max value. Second,pooling on 4 local subsections ,each a quarter the size of  the feature map  producing 4  max values,each for one subsection. After spatial pyramid pooling, a vector of length 5 is generated.The  vectors of  each feature map are then flattened to form the final feature vector of length 1360($5\times272$).From Figure3(b) we can see that the size of the final feature vector depends only on the number of subsections of each feature map as well as  the number of filters of the preceding convolutional layer  so that the spatial pyramid pooling layer can produce a feature vector of fixed length given an arbitrarily-sized input.The advantage of  using spatial pyramid pooling layer is that we can train our network with  arbitrarily-sized QRS complex and sampling rate which increase scale-invariance and reduce over-fitting\cite{Kaiming2014Spatial}.Besides, the trained network can be applied to a wider range of ECG data with different length and sampling rate collected using different devices.
Similar to the spatial pyramid pooling is the global max pooling inspired from global average pooling in\cite{Network}.It replaces the final fully connected layer by performing a max-pooling operation on the entire feature map to accept arbitrarily-sized input.As can be seen,global max pooling is in fact a special case of  spatial pyramid pooling when max pooling is only applied to the entire feature map.

\begin{figure*}[t]
\centering
\includegraphics[width=1\linewidth]{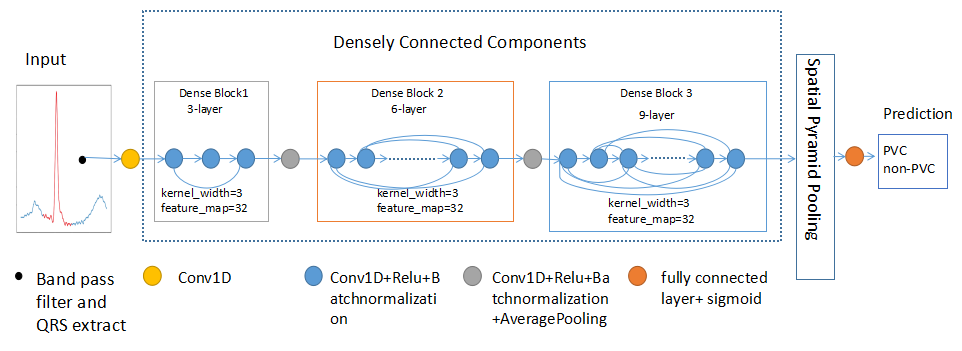}\vspace{1mm}
\caption{ The proposed  diagram for PVC detection.}
\label{fig:arch}\vspace{-2mm}
\end{figure*}

\subsection{Weighted binary cross-entropy loss}

As can be seen in Table 1,the number of PVC and non-PVC beats is highly imbalanced. For each training epoch, non-PVC samples contribute much more than PVC samples to updating network parameters which will severely undermines the performance of the network to detect PVC.To address the problem,a weighted binary cross-entropy loss function is adopted as in \textbf{Eq.2}:

\begin{equation}
{\l=\sum_{x\in X}\eta _{\iota (x)}[y^{true}lny^{out}+(1-y^{true})ln(1-y^{out}))]} 
\end{equation}

Where $\iota(x)$ is the target label for sample $x$ and $|X^{\iota(x)}|$ is the number of  samples belong to target  $\iota(x)$  in training batch X. $y^{out}$ and $y^{true}$ are the model output and groundtruth respectively.$\eta _{\iota (x)}$ is the weight coefficient:

\begin{equation}
{\eta _{\iota (x)}=1-\frac{|X^{\iota (x)}|}{X}} 
\end{equation}

\textbf{Eq.2} means that  classes with fewer samples are given a larger weight so that PVC samples contribute to the loss equally as non-PVC samples in each training epoch. $-l$ is minimized in training.

\begin{figure}[h]
\centering
\includegraphics[width=.99\linewidth]{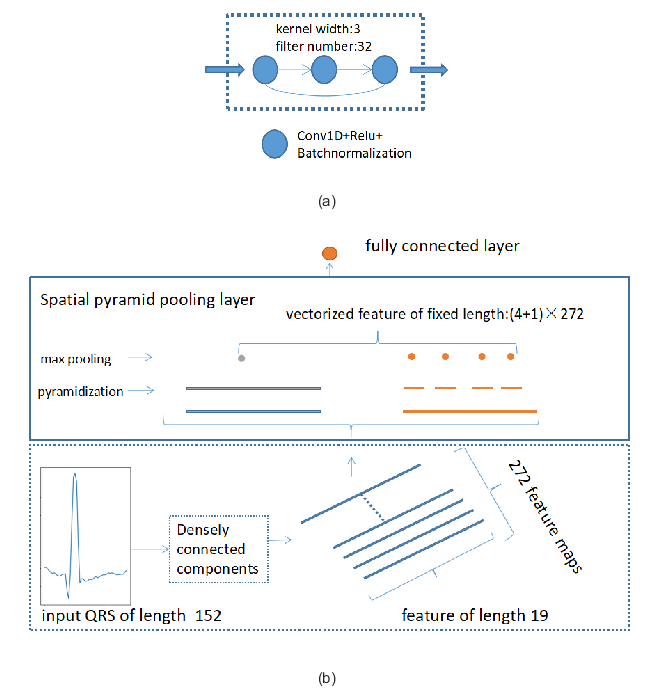}
\caption{(a)A 3-layer dense block with no bottleneck.The filter number of each convolutional layer is 32 and the kernel width is 3.(b)A two-level spatial pyramid pooling layer used in our network.}
\end{figure}

\section{Experiments and Results}
Three experiments are conducted in this section.First,to have a comparison with the results of  other published  methods, the proposed network is  trained and tested using only DS1. The splitting strategy for training set and testing set  is the same as in\cite{Zarei,Zhou2017Premature}.Second,to demonstrate how training on multiple databases affect the performance of the network compared to training using a single database,all the training set of the five databases described in Table 1 are used for training in our second experiment.Testing is also carried out  on the test sets of DS1-DS5.The third experiment we conduct is a simple ablation study to remove the weighted loss function and spatial pyramid pooling layer for training.This experiment is to demonstrate that weighted loss function and a spatial pyramid pooling layer is necessary to guarantee good performance.The experiment is carried out on DS1.Metrics used in our experiments to quantify  the performance of PVC detection are accuracy,sensitivity, specificity,positive predictive value(PPV), and Youden’s index $\gamma$.

\subsection{Training details}
For all the experiments, the training strategy and model configuration are kept the same.The network is implemented in Keras and Tensorflow on a machine with  32GB memory and Core i7-6700 CPU(8 cores with 16 threads) and a 8Gb Nvidia Quadro M4000 GPU.The computation power of the machine is fully exploited to accelerate training. The initial learning rate is set to 0.001 and decreases by 5 percent every 100 epochs.The batch size is set to 100.20 percent of the training set is used for validation and training stops when validation loss stops to decrease.Adam optimizer is used for updating network parameters.  For all experiments,the input are normalized to the range[-1,1] before feeding into the network. The network contains approximately 360000 trainable parameters.
Our codes and checkpoint will be available upon acceptance\footnote{https://www.github.com/Eric-THU/PVC-Detection}
\subsection{Training and testing on DS1}
In our first experiment,the network is trained and tested on DS1.In this case, our network accepts single input size of 150 both in training and testing.The splitting strategy for training and testing set  in Table 1 is the same with that of \cite{Zarei,Zhou2017Premature} for comparison purpose.Results are shown in Table 2.

\begin{table*}[h]
 \caption{Comparison with other published methods on DS1 }
 \label{sample-table}
 \centering
 \begin{tabular}{lllccc}
   \hline
    methods          &Acc        &Sensitivity        &Specificity        &PPV           &$\gamma$   \\
   \hline
  Zarei\cite{Zarei}            &98.77              & 96.12            &  98.96       &86.48          & 95.08 \\
  Fei-yan Zhou\cite{Zhou2017Premature}    & 99.41              & 97.59            &  99.54      & 93.55          & 97.13  \\
  Proposed           &99.26                &97.37             & 99.39       &92.23          & 96.76 \\
   \hline
 \end{tabular}
\end{table*}

As can be seen from table  2,our proposed method  produces results   comparable to current state-of-the-art method by\cite{Zhou2017Premature}on the test set of DS1 and the network architecture and training procedure of our method is much less complicated  than theirs.

\subsection{Training and testing on DS1-DS5}
Previous deep learning based methods for PVC detection require that the length of input QRS complex be fixed.Training and testing are conducted  on the same database splitting into training set and testing set where the QRS complexes are of the same length and sampling rate.As is detailed in Methodology,our network can take as input QRS complex of arbitrary length.Therefore, we can use multiple databases for training and testing .In this experiment,the network is trained using all the training sets of DS1~DS5 detailed in Table 1  and tested on all  the test sets.Altogether,there are 4 sizes for input which is 150,52,108 and 257.
Training on multi-size QRS complexes is a tricky task.We train our network iteratively  from one database to another.In our study,the network is trained 20 epochs on one database before switching to another.When the validation loss while training on a database stops  decreasing,the database is discarded and does not involve in the next training iteration.This is iterated until all databases are discarded.Testing results are shown in Table 3.

\begin{table*}[h]
 \caption{Testing results on DS1-DS5}
 \label{sample-table}
 \centering
 \begin{tabular}{lllccc}
   \hline
    Database          &Acc        &Sensitivity        &Specificity        &PPV           &$\gamma$   \\
   \hline
    DS1                 &99.43             &98.19             &99. 52            &93.95       & 97.71 \\
    DS2                 &93.92             &89.13              &95.49            &86.57       & 84.62  \\
    DS3                 &95.92              &80.00              &96.00             &  /       & 76.00  \\
    DS4                 &94.34             &93.26             & 96.73            & 98.44      & 89.99  \\ 
    DS5                 &94.08             &91.36              &94.57            & 75.17      & 85.93  \\
    Overall             &95.58              &92.68              &97.05            & 94.09      & 89.73 \\ 
   \hline
 \end{tabular}
\end{table*}

By comparing Table 2-3, we can see that the performance on DS1 is  improved by training on multi-size input (Table 2 row 3 and Table 3 row 1).What’s worthy of note in this experiment is that the improvement on DS1 is not merely because of  more training data.A sensible explanation should be that by training on multiple databases with multi-size ECG data  sampled at different rates  and from more unique individuals, the network can learn more generalized and abstract features of PVC  and non-PVC.In other words, the generalization capability of the network can be improved through training on a database of greater diversity compared to training using a single database.
The network also achieves satisfactory results on the test sets of DS2~DS5.Note that ,in DS3,4 out of 5 PVC beats are correctly identified. 
Altogether,our network achieves an overall accuracy of 95.58 percent with sensitivity and specificity being $92.68\%$ and $97.05\%$ respectively on a mixed database containing 62200 PVC samples and  122771 non-PVC samples.

\subsection{Ablation study}

A simple ablation study is conducted to quantitatively evaluate how weighted loss function and the spatial pyramid layer affect the network performance.In this experiment,three simplifications of the proposed network is trained and tested on DS1.First, a standard 20-layer CNN without dense connection components and spatial pyramid pooling layer.The first 19 layers are convolutional layers followed by max-pooling and the last layer is the single-node fully connected layer with sigmoid activation function for binary classification. 
Second ,the weight coefficient of the loss function is removed.Third,the spatial pyramid pooling layer is replaced by a global max pooling(GMP)layer.For comparison purpose, all the networks have approximately the same number of trainable parameters.Results is shown in Table 4.

\begin{table*}[h]
 \caption{Ablation study results on DS1 }
 \label{sample-table}
 \centering
 \begin{tabular}{lllccc}
   \hline
    methods          &Acc        &Sensitivity        &Specificity        &PPV           & $\gamma$  \\
   \hline
Standard 20-layer CNN            & 94.08               & 80.10                 & 95.05             & 53.03         &75.15 \\
DenseNet+GMP+weighted loss         & 97.50               & 92.43                 & 97.85             & 75.21        & 90.28   \\
DenseNet+Spp+unweighted loss       &  96.47             &   83.26                 & 97.63             & 70.25        & 80.89   \\
DenseNet+Spp+Weighted loss           & 99.26             &  97.37                 & 99.39             &  92.23        & 96.76  \\
   \hline
 \end{tabular}
\end{table*}

By analyzing the results in Table 4, we can find that both dense connection components and spatial pyramid pooling helps improve PVC detection performance.More specifically,sensitivity is severely degenerated when the weight coefficient in Eq.2 is removed which is in accordance with our assumption that the contribution of PVC samples to the loss is overwhelmed when non-PVC samples far outnumber PVC samples.This causes the network fail to learn adequately the features of PVC leading to a poor PVC distinguishability.To further illustrate,we visualize the histogram of the output of the final fully connected layer for each training step in Figure 4.In Figure 4 (a), the network output is totally biased towards non-PVC samples (output value well below 0.5) while Figure 4 (a),the network is able to distinguish between PVC samples and non-PVC samples with good decision margin.
On the other hand,more levels of pyramid in the spatial pyramid pooling layer improve overall performance.Compared with the 2-level pyramid {$1\times1$,$2\times2$} we use in our network,we see a drop in accuracy,sensitivity and specificity when it’s replaced by global max pooling which can be seen as a 1-level pyramid{$1\times1$}.This phenomenon can be explained that by using more levels of pyramid,the network can become more robust to the variations of QRS complex due to individual difference.However,it remains to be experimented whether the  PVC detection performance can keep improving as the level of pyramid in the spatial pyramid pooling layer goes higher. The standard 20-layer CNN without dense connection components and spatial pyramid pooling performs the worst.What’s interesting is that  the network converges at a high accuracy (near 100 percent) and a low loss (near 0)  at the end of training but can not generalize well in testing set showing that it’s over-fitted and has poor generalization capability.

\begin{figure}[h]
\centering
\includegraphics[width=.99\linewidth]{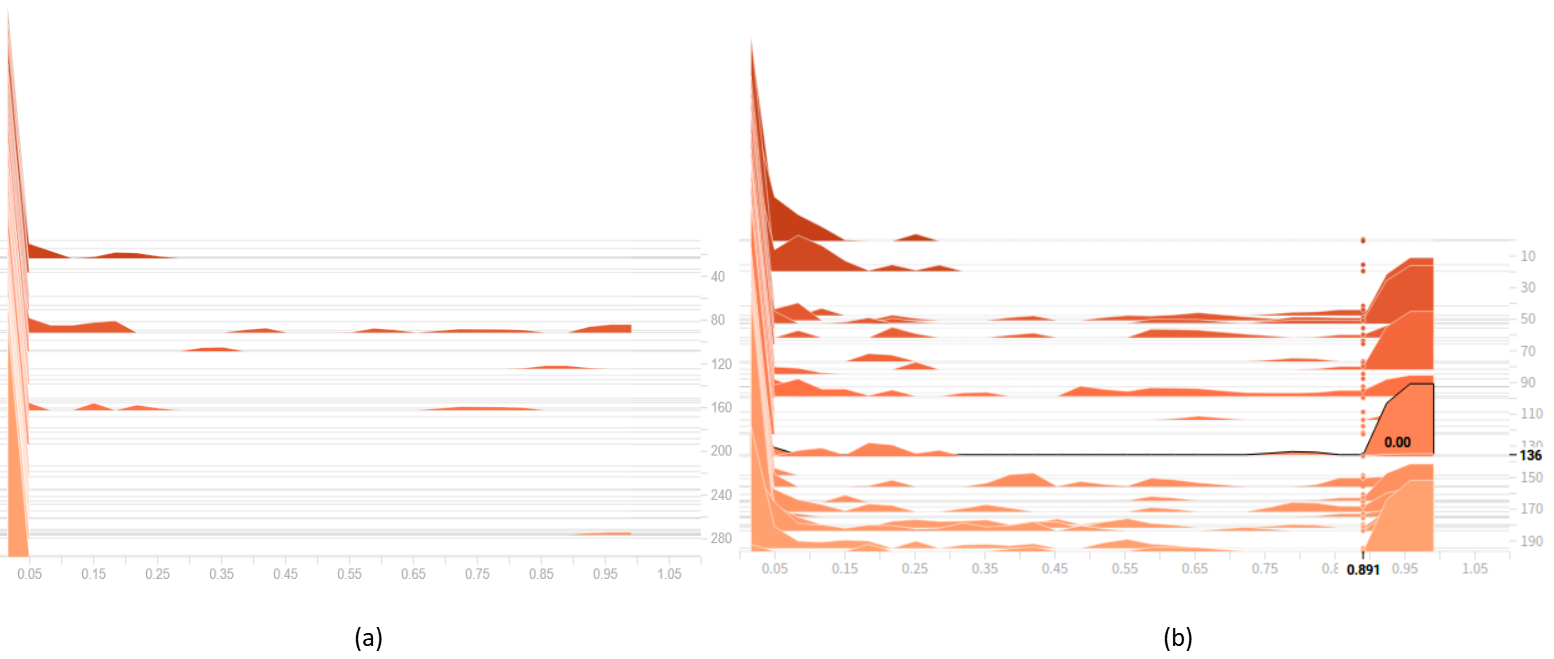}
\caption{The histogram of the output of the final fully connection layer for each training step without(a)  and with(b) weighted loss function.}
\end{figure}

\subsection{PVC detection with focal loss}
In 3.3, a weighted binary cross-entropy loss function is adopted to address the issue of high class imbalance. By introducing the weight coefficient to the loss function, the loss generated by PVC samples can avoid being overwhelmed by the loss from non-PVC samples. In our experiments we also find that it’s an easy task to achieve a high specificity even with an ordinary network while a well-designed network is required to guarantee  high sensitivity. In other words, the non-PVC samples which are dominant in number are easy negative samples while the PVC samples belong to hard positive samples , quoting the term from\cite{Lin2017Focal}. To address the problem, it’s natural to think of  paying less attention to these easy negative samples while focusing more on the hard positive samples in the loss function for each training epoch. Inspired by \cite{Lin2017Focal}, we also experiment using a modulating factor $(1-p_{t})^{\gamma }$ together with the weight coefficient  $\eta _{\iota (x)}$ in \textbf{Eq.2}.

\begin{equation}
{\ p_{t}=\left\{\begin{matrix} y^{out},(y^{true}=1) \\ 1- y^{out},(y^{true}=0)\\ \end{matrix}\right.} 
\end{equation}

By rewriting \textbf{Eq.2}. we get the focal loss function as in \textbf{Eq.4}:
\begin{equation}
{\l_{focal}=\sum_{x\in X}\eta _{\iota (x)}(1-p_{t})^{\gamma }ln(p_{t}),\gamma =3} 
\end{equation}

It’s apparent from \textbf{Eq.3} and \textbf{Eq.4} that if a QRS complex is correctly classified as non-PVC , $y^{out}$ can be small.In this situation, $p_{t}=1-y^{out}$ so that $p_{t}$ is large  and the modulating factor $(1-p_{t})^{\gamma }$ in \textbf{Eq.4} is small. On the contrary, if a PVC QRS complex is misclassified as non-PVC,the modulating factor will be large. By using \textbf{Eq.4} as loss function, the easily misclassified hard positive samples (PVC samples) contribute more to the loss than the easy negative samples so that the network can be more sensitive to PVC. Figure.5 shows the comparison between weighted loss and focal loss under the framework of denseNet with spatial pyramid pooling (spp) layer.

\begin{figure}[h]
\centering
\includegraphics[width=1\linewidth]{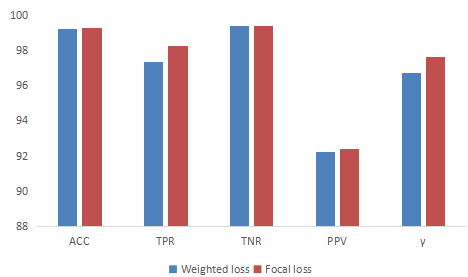}
\caption{Comparison between weighted loss and focal loss with regard to sensitivity,specificity,accuracy,positive predictive value and Youden's index.}
\end{figure}

We can see that sensitivity(true positive rate TPR) is improved which means that the network becomes more sensitive to hard positive samples by training with focal loss.

\subsection{Looking deeper into  deep  convolutional neural network  for PVC detection}
 A human doctor can easily distinguish PVC beat from non-PVC  beat through the obvious morphology features of PVC  depicted in electrocardiogram (ECG) caused by abnormal electrical event of ventricles .As can be seen in Figure 1 top row,the QRS complex of a PVC beat is wider compared to normal beat and depicts abnormality.And the direction of the S-T segment and T wave is opposite to the QRS complex.These are the most apparent morphology features to which doctors are paying attention while  detecting PVC. The morphology of a PVC QRS complex can have numberous variations due to individuals difference and different ECG devices.However,experienced doctors are still able to correctly identify PVC regardless of these variations.This is because they have learned the general features of  PVC and  rely on those features for making diagnosis.Contrary to general features are patient-specific features which are unique to only a small number of individuals. If doctors rely on patient-specific features to make diagnosis, there’s big chance of missed diagnosis.In other words,the sensitivity is expected to be low.The same is true to deep convolutional neural network when it’s used to detect PVC.  
To generalize well, the features learned by  convolutional layers should be general and independent from specific individuals and ECG sampling rates.One way to know whether features learned by a trained network is general or not is to visualize what part of  the input  QRS complex the network is focusing on in order to classify it as PVC.
Using the techniques described in \cite{Zeiler2013Visualizing}, attention maps are generated for some PVC inputs .
The one-dimensional attention map is of the same length as the input QRS complex. The intensity value  represents 
how much  the network is focusing on the part  to  generate the prediction. As can be seen in Figure 6 (a~f),the highest attention intensity values are around the location of S wave.This shows that the network thinks an reverse S-T segment  should be the criterion to classify an QRS complex as PVC which is in accordance with doctors’ experience.On the contrary,it can be imagined that if the attention intensity values are distributed evenly across the entire QRS complex or  peaks at irrelevant  locations,the network is expected to have  poor generalization capability.

\begin{figure*}[h]
\centering
\includegraphics[width=.99\linewidth]{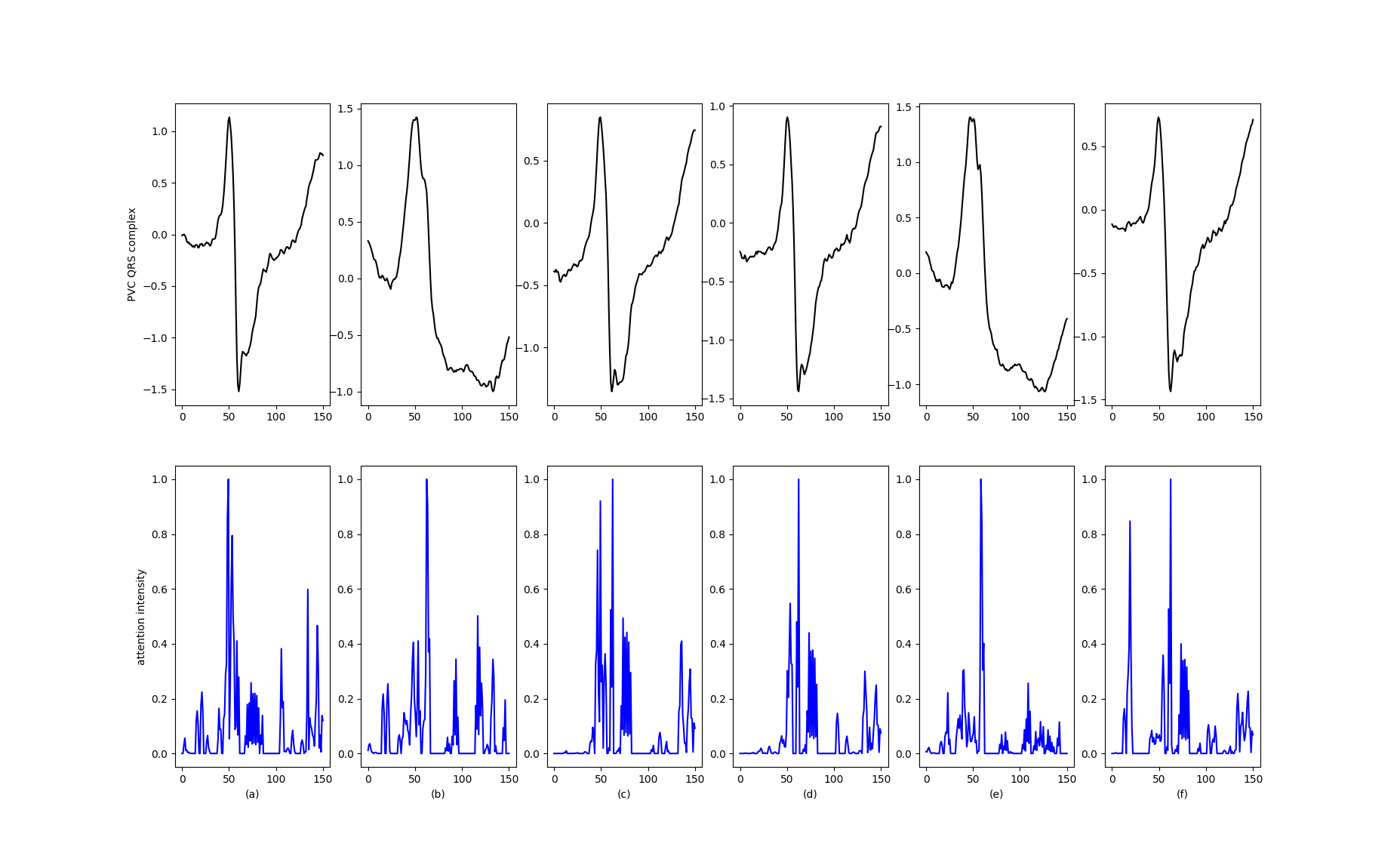}
\caption{Examples of input QRS complex(top row) and the corresponding attention map(bottom row).}
\end{figure*}

\section{Conclusion and discussion}
We propose in our study an automated method combining densely connected convolutional neural network and a 2-level spatial pyramid pooling layer for PVC detection.The proposed network can be trained using multiple databases where the QRS complexes  have different length and sampling rates.Our method achieves comparable results on the MIT-BIH arrhythmia database  with current state-of -the -art PVC detection network based on deep learning.And the architecture of  our proposed network is much  more straightforward and easier to implement. 
We demonstrate that training using QRS complex of varied input size and sampling rates helps improve overall performance compared using only a single database. The generalization capability  of our method is validated on 4 more databases other than the MIT-BIH arrhythmia database and can be applied to a wider range of ECG data collected  by different devices.It remains to be experimented  whether the network can maintain good performance when testing on ECG data with sampling rate not included in training set.
From the point of view of object detection in 2-dimensioanl images, a bounding box is predicted around the targeted object in an image.For  the detection of PVC ,the prediction should be a bounding box in one-dimensional  ECG time-series  which is equivalent to segmenting the QRS complex from an ECG cycle.An one-dimensional DenseUNet  which is trained end-to-end  is proposed  to perform the task.As can be seen in Figure 7, the max pooling  operations between two transition layers in Figure 2 are replaced by down-sampling and up-sampling operation respectively as in UNet\cite{Ronneberger2015U}. Different from traditional PVC detection methods which performs a binary classification  towards a QRS complex, the proposed one-dimensional DenseUNet performs a point-wise classification on a ECG cycle.As  is illustrated in Figure.5, the network takes as input an ECG cycle and outputs a sequence of the same size as input that indicate the QRS complex of PVC.The size of  input can be arbitrary since there is no fully connected layer in the network.The performance of  detecting PVC in the way as object detection in 2-dimensional images needs further experiments.Besides,the proposed network can be easily extended to the detection of other arrhythmia such as atrial fibrillation,premature atrial contraction(PAC) ,etc with a suitable database for training.

\begin{figure*}[t]
\centering
\includegraphics[width=1\linewidth]{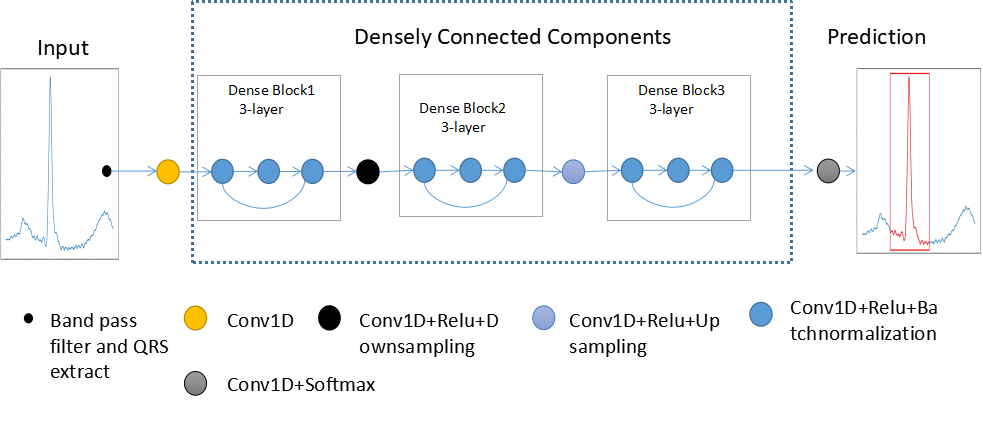}\vspace{1mm}
\caption{A  deep learning network  based on 1-dimensional denseUNet  for PVC detection.}
\label{fig:arch}\vspace{-2mm}
\end{figure*}

{\small
\bibliographystyle{ieee}
\bibliography{egbib}
}

\end{document}